\documentclass[letterpaper, 10 pt, conference]{ieeeconf}  

\IEEEoverridecommandlockouts                              

\usepackage{cite}
\usepackage{amsmath,amssymb,amsfonts}
\usepackage{algorithmic}
\usepackage{graphicx}
\usepackage{textcomp}
\usepackage{xcolor}
\usepackage{booktabs} 
\usepackage{multirow}

\def\BibTeX{{\rm B\kern-.05em{\sc i\kern-.025em b}\kern-.08em
    T\kern-.1667em\lower.7ex\hbox{E}\kern-.125emX}}
\begin{document}
\title{\LARGE \bf Intent-guided Adaptive Stiffening for Visuomotor Policies under Unknown Resistance}
\author{Francesco Cufino$^{1}$, Carmela Mariniello$^{1}$, Mario Selvaggio$^{1}$,
        Andreas Sochopoulos$^{2}$, Fabio Ruggiero$^{1}$
\thanks{$^{1}$PRISMA Lab, Department of Electrical Engineering and Information
Technology, University of Naples Federico II, Naples, Italy.
{\tt\small francesco.cufino@unina.it}}
\thanks{$^{2}$The University of Edinburgh, Edinburgh, UK.}
\thanks{Submitted to ICRA 2027}
}
%
\maketitle
\begin{abstract}
Manipulation tasks such as unscrewing a cap or sliding a part in contact with a surface require imposing force along a high-friction direction: the robot must overcome a resistance rather than yield to it. 
Generative visuomotor policies predict the motion for such tasks but not the authority needed to realize it, and that authority is bounded by the impedance controller beneath them: when the resistance exceeds what a compliant controller can exert under a bounded tracking error, the task is unattainable however accurate the prediction is.
The required amount cannot be anticipated either, since it depends on a contact state the observation does not reveal. 
In this paper, we propose a residual stiffening layer placed underneath an unmodified policy. 
Resistance is recognized, without force sensing, from the mismatch between the motion the policy commands and the motion the robot realizes, and a four-mode supervisor raises the stiffness along a single direction read off the policy's own action chunk, retains it while the task progresses and releases it afterwards. 
The layer needs no compliance labels, no retraining and no re-collected demonstrations. 
On two real tasks, unscrewing a bottle cap and pushing a T-nut along a T-slot groove against non-homogeneous friction, evaluated at six resistance levels of which two were withheld from the demonstrations, the layer raises the success rate from $52\%$ to $90\%$ while producing $61\%$ less peak off-axis force than a uniformly stiff controller at the same resistance.
\end{abstract}

%
%
\section{Introduction}
Manipulation remains a central challenge in robotics: physical interaction with the environment demands careful handling of contacts, friction, dynamics, and forces. Recent progress has come from imitation learning, which leverages human demonstrations to solve complex manipulation tasks in unstructured settings. A key difficulty in this setting is multimodality, the fact that a task often admits several valid solutions, which generative approaches such as diffusion policies and flow matching have proven effective at capturing~\cite{chi2025diffusion, lipman2022flow}.
Most methods in this class, however, are largely position-centric: they predict where the end-effector should go, leaving the interaction forces that arise along the way to whatever controller executes the command~\cite{hou2025adaptive, VIDP, huang2026unireflex}. 

Two classes of tasks expose the limits of this separation. First, tasks in which the motion is kinematically constrained, so that a small positional error is converted directly into a large interaction force. Second, tasks in which the quantity that determines the required effort is not observable at all: how tightly a cap is screwed, or how much friction resists a sliding part, is invisible to a camera and absent from the robot's proprioception. In both cases the demonstrated trajectory alone does not determine the outcome, and the robot must instead regulate how firmly it commits to that trajectory along the direction required by the task, regulating the interaction force either with direct force control, or indirectly, with impedance control.
%
%
In the latter, the stiffness exhibited via control can be made anisotropic, so that authority is granted along the direction the task demands while the remaining ones stay compliant and safe to contact~\cite{hogan1984impedance, ott2008, simonivc2026plug}.

\begin{figure}
    \vspace{0.3cm}
    \centering
    \includegraphics[trim=0 2cm 0 2cm, clip, width=1\linewidth]{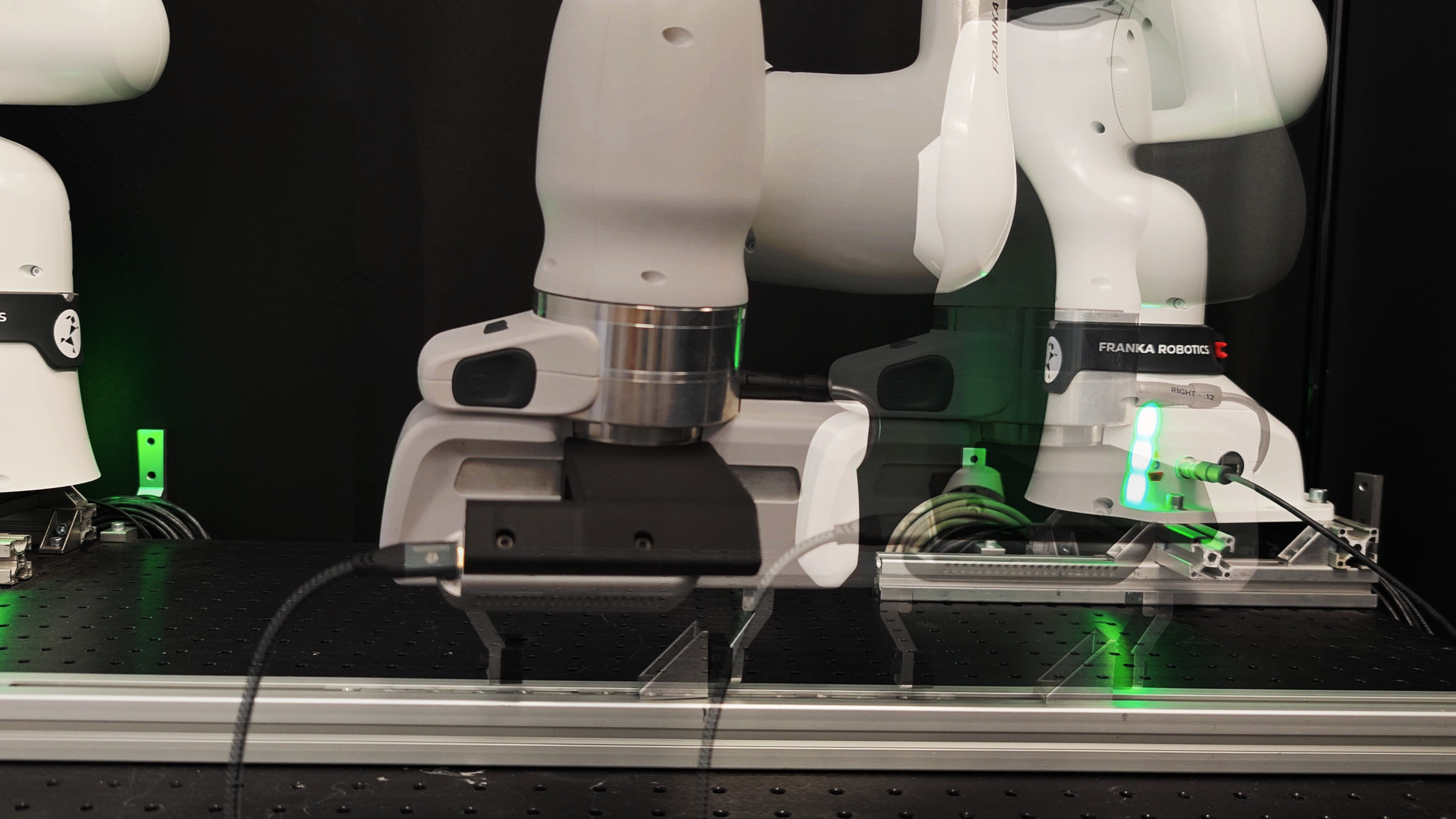}
    \caption{Example task demonstrated by our method. 
    A robot leverages a learned visuomotor policy to slide an angle bracket mounted onto a T-nut through a high-friction aluminum profile groove. 
    Our adaptive method dynamically regulates robot stiffness to overcome non-homogeneous resistance, extracting the intended stiffening direction directly from policy action predictions.}
    \label{fig:push_task}
    \vspace{-0.4cm}
\end{figure}

Several methods adapt the compliance at runtime to improve performance in contact-rich tasks, and this has attracted recent attention in imitation learning. They share a common assumption, however: that a larger interaction force calls for more compliance, since yielding is what prevents jamming and keeps contact forces bounded~\cite{hou2025adaptive, VIDP, aburub2026learning, kamijo2024learning}.
This is inverted in an important family of tasks, those requiring force imposition along a high-friction direction, such as unscrewing a cap, sliding a stiff drawer open, or pushing a part against friction (see Fig.~\ref{fig:push_task}). Here the robot must overcome the resistance rather than yield to it, which demands more control authority along the task direction, while compliance everywhere else remains just as necessary.
In this work we focus on this family of tasks, where the resistance to be
overcome is unknown at execution time and may exceed anything encountered
during demonstration.

\subsection{Motivation and objectives}

Tasks of this family must be executed under impedance control. 
The motion is kinematically constrained, so a positional error of a few millimetres or radians is converted into a large reaction force or torque that might damage the object or the robot~\cite{hogan1984impedance, ott2008}. 
Errors of that
magnitude are unavoidable with a visuomotor policy, making compliance selection and interaction force bounding a critical step.
%
Since that force is the stiffness times the tracking error, letting the error grow unbounded must be avoided: the virtual spring stores energy quadratically in the displacement, and a distant set-point may produce a large acceleration when the resistance is suddenly overcome.
Hence, the controller should keep the equilibrium offset bounded, which, at low stiffness, implies a bounded force. 
The tension is that the same compliance must both be low enough to be safe on contact and high enough to overcome an unexpected resistance, whose magnitude cannot be anticipated from an observation that does not reveal the contact state.

Our objective is to raise the success rate of a visuomotor policy executed on an impedance-controlled robot in exactly these conditions, where the bottleneck is the force authority available to execute what it commands. 
To realize this, we propose a stiffening adaptation layer placed underneath the policy and
transparent to it. 
What the policy provides is the task direction: the mean of its action
chunk over the upcoming horizon, which we call the policy \emph{intent}, and it is along that direction, and only along it, that authority is granted. 
Independent operation is insufficient; the classical adaptation law cannot allocate control authority, and the policy cannot quantify the required stiffness. 
We therefore report the success rate of their combined framework.

The contributions of this work are here summarized.
\begin{itemize}
  \item We characterize an execution-side failure mode of generative visuomotor policies in force-imposition tasks: the policy commands appropriate motion,
  but a compliant impedance controller cannot overcome the encountered resistance under a bounded tracking error.

  \item We propose a residual stiffening layer coupled to the policy through the only signal it needs, the action chunk. Resistance is detected from the mismatch between commanded and realized motion, and the stiffness is raised
  along the direction of the policy's own intent, with no retraining, no
  compliance labels and no force sensing.

  \item We analyze the energetic behaviour of the layer and design it so that neither raising the stiffness nor reorienting it injects energy into the
  interaction.

  \item We validate the approach on two real tasks, each demonstrated at four resistance levels and evaluated at six. The layer raises the success rate on both, including at the two withheld levels, while producing substantially less
  off-axis force than a uniformly stiff controller.
\end{itemize}

\section{Related works}

Variable impedance control has a long history~\cite{abudakka2020review}, from
model-based stiffness computation to reinforcement
learning~\cite{martinmartin2019variable} and physical human-robot interaction,
where the objective is to yield to the partner. We focus on methods that
produce a stiffness signal for the autonomous execution of a learned policy,
and organize them by where that signal comes from.

\paragraph{Stiffness derived from demonstrations}
The dominant approach derives compliance from the demonstration data itself,
either as a profile fitted offline or as an output supervised during training.
The statistical line maps the variability of the demonstrated motion onto
stiffness, from the original covariance-based
formulation~\cite{calinon2010learning} to its recent extension to spatially
diverse demonstrations~\cite{VIDP}: no force sensing is needed, but the profile
is fixed at training time and the mapping is monotone in one direction only,
with high variance always yielding low stiffness. The supervised line predicts stiffness alongside the action, from labels
produced by an operator toggling between hand-tuned
levels~\cite{aburub2026learning}, from a geometric rule that lowers stiffness
along the measured force direction~\cite{hou2025adaptive}, or from a
feedforward force target learned jointly with the pose~\cite{CR-VLA}. A third line attaches the regulation to a policy left unmodified, which is
closest to our architecture: a reflex network trained on latent representations
of a frozen backbone predicts anisotropic stiffness
directions~\cite{huang2026unireflex}, and an energy-based estimator derives a
directional adaptation from a contact-consistent equilibrium
trajectory~\cite{geiger2025diffusion}. What all of these share is that
the attainable authority is encoded at design or demonstration time, and is
therefore bounded by what the demonstrations contain; those that adapt online
do so from force feedback and downwards from a high baseline, which is the
opposite of what force imposition requires.

\paragraph{Model-free impedance adaptation}
The classical relatives of our work adapt impedance online from tracking error
alone. The biomimetic controller in~\cite{yang2011humanlike}, derived from the
minimization of instability, motion error and effort, adapts feedforward force
and stiffness concurrently and, notably, requires no force measurement; it is
grounded in the observation that humans stiffen selectively to stabilize
unstable dynamics~\cite{burdet2001cns}, and was later extended to contact
tooling~\cite{li2018force}. These controllers share our premise, that authority
can be acquired at run time without sensing force, but they carry no notion of a
task direction: the adaptation is isotropic or expressed in a fixed frame, so
the acquired stiffness is not confined to where it is needed. Their update law
also includes a continuously active forgetting term, appropriate for iterative
convergence over repeated trials but, in a breakaway event, releasing the
acquired stiffness as soon as motion resumes. Our supervisor supplies what is
missing in this setting: a direction, read off the policy, and a hold phase that
retains the stiffness until the resistance is genuinely past.

\paragraph{Stiffening under load}
A small number of works do increase stiffness with interaction force.
In~\cite{guan2021improvement, xia2025bio} stiffness is derived from human muscle
activation recorded during teleoperated demonstrations, so that when the
operator stiffens their arm the robot stiffens accordingly; this inverts the
usual compliance rule, but by copying a biological signal rather than deriving
it from the task. In~\cite{abu2018force} a force-to-stiffness map is learned by
Gaussian mixture regression from kinesthetic demonstrations, unable to
generalize beyond the demonstrated force range. In both cases stiffness remains
an imitated label rather than a quantity derived, at run time, from what the
task actually requires. Across all the lines reviewed here, stiffness is either
fixed at design time, imitated from a demonstrator, or adapted downwards to keep
contact forces bounded, and none is evaluated on generalization to unseen,
higher-resistance conditions.

\section{Methodology}
\label{sec:method}

In this section we describe the pipeline proposed in this work: a generative
policy produces action chunks in twist space, and a layer designed for
force-imposition tasks, placed underneath it, decides how much authority is
committed to realizing them. Doing so requires detecting that the robot is
being resisted without measuring force, and discovering how much authority the
resistance demands when that quantity is unknown a priori. The policy is treated as a black box: it observes no forces, predicts no stiffness, and is not
retrained.

The first two subsections recall what the pipeline builds on, the policy
interface and the baseline impedance controller; the remaining ones describe
the contribution. The interface between policy and layer is a single signal,
the action chunk, which carries both the commands the controller executes and
the intent the layer reads.

\subsection{Policy interface and instantiation}
\label{sec:policy}

We assume access to a policy $\pi_\theta$ that takes as input an observation $o\in\mathcal{O}$, which can be of any modality (in this work, we employ specifically policies that accept RGB
images and proprioception) , and outputs an action chunk
$A\in\mathbb{R}^{H\times 6}$, i.e., a sequence of $H$ twists covering a finite
horizon. Overall, we impose no assumptions on the nature of the base policy. 

However, we use flow-based visuomotor policies, trained with imitation learning, as base policies throughout the paper. We train $\pi_\theta$ from a demonstration dataset
$\mathcal{D}=\{(o^{(i)},A^{(i)})\}_{i=1}^{n}$ and frame imitation as
conditional generative modeling: each $A^{(i)}$ is regarded as a draw from a
target distribution $p_A(A\mid o)$ evaluated at $o=o^{(i)}$.

Concretely, we look for a neural vector field
$v_\theta:[0,1]\times\mathbb{R}^{H\times 6}\times\mathcal{O}\to\mathbb{R}^{H\times 6}$
such that, integrating $\mathrm{d}A_s = v_\theta(s, A_s\mid o)\,\mathrm{d}s$
over $s\in[0,1]$ from $A_0\sim p_0=\mathcal{N}(0,I)$, the endpoint $A_1$ is
distributed according to $p_A(\cdot\mid o)$. We fit $v_\theta$ with
conditional flow matching (CFM)~\cite{lipman2022flow}. CFM requires a coupling
$q(A_0,A_1\mid o)$~\cite{tong2024improving, sochopoulos2025fast} with marginals $p_0$ and $p_A(\cdot\mid o)$, which we use to sample noise-action pairs, together with a path joining
each pair. We use the straight line $A_s=(1-s)A_0+sA_1$, which is generated by
the constant conditional velocity $u(s,A_s\mid A_0,A_1)=A_1-A_0$. The network
is trained using the following objective:
\begin{equation}
  \mathcal{L}_{\mathrm{CFM}}(\theta) :=
  \mathop{\mathbb{E}}_{\substack{s\sim\mathcal{U}([0,1]),\\ o\sim\mathcal{D}\\
  (A_0,A_1)\sim q}}
  \big\| v_\theta\big(s,\,A_s \mid o\big) - (A_1 - A_0) \big\|^2 .
\end{equation}
The vector field minimizing this loss carries $p_0$ onto $p_A(\cdot\mid o)$ at
$s=1$. A flow-based $\pi_\theta$ trained with CFM is our
choice of implementation rather than a requirement: the proposed adaptation layer only reads the
chunk and its duration, so it is compatible with any policy that outputs a
sequence of twists over a horizon.
Besides serving as a motion command, the chunk plays a second role in our
method. It encodes the motion the policy plans to perform over the upcoming
horizon, and Sec.~\ref{sec:direction} extracts from this planned motion the
direction in which the layer must grant authority. The chunk does not,
however, specify how the robot physically interacts with its environment
while executing that motion. This is governed by the low-level controller,
introduced below.

\subsection{Impedance control}
\label{sec:impedance}

Let $q\in\mathbb R^{n}$ be the robot joint positions, with $n$ the number of joints, $p_e\in\mathbb R^3$ and
$R_e\in SO(3)$ the end-effector position and orientation, $J\in\mathbb R^{6\times n}$ the world
Jacobian, $M\in\mathbb R^{n\times n}$ the joint inertia matrix and
$\Lambda=(JM^{-1}J^{\!\top})^{-1}$ the Cartesian inertia. One twist per
control period is streamed from the chunk and integrated into a virtual
equilibrium pose $(p_d,R_d)$, giving the position error
$\tilde p = p_d-p_e\in\mathbb R^3$ and the orientation error
$e_O = \tfrac12\sum_{j} r_{e,j}\times r_{d,j} \in \mathbb{R}^3$, with
$r_{e,j}$ and $r_{d,j}$ the columns of $R_e$ and $R_d$~\cite{ott2008}. The commanded torque is the
operational-space law~\cite{khatib1987unified,nakanishi2008operational}
\begin{equation}
\tau_0 = J^{\!\top}\Lambda\big(\ddot x_r - \dot J\dot q\big) + C
       + \big(I - J^{\!\top}\bar J^{\!\top}\big)\tau_{\mathrm{null}},
\label{eq:osc}
\end{equation}
where $\ddot x_r \in \mathbb{R}^6$ is a PD term on the complete error vector
$\tilde{x} = (\tilde p^{\!\top},\, e_O^{\!\top})^{\!\top} \in \mathbb{R}^6$
with gains $K_a, D_a \in \mathbb{R}^{6\times6}$ acting in the acceleration
domain, in $\mathrm{s}^{-2}$ and $\mathrm{s}^{-1}$ respectively, while $C \in \mathbb{R}^{n \times n}$ collects the Centrifugal/Coriolis terms, gravity is compensated in
firmware, and $\tau_{\mathrm{null}}$ is a nullspace term acting through the
dynamically consistent pseudoinverse $\bar J$~\cite{khatib1987unified}.

A compliant robot exerts a force equal to its stiffness times its tracking
error, so both quantities need an upper bound for safety. The bound on the error
is the more critical, because the elastic energy stored by the virtual spring,
$\tfrac12\tilde x^{\!\top}K_c\tilde x$, grows quadratically in the displacement,
and here the equilibrium pose is an open-loop integrator that keeps drifting
away from a stalled end-effector. We therefore saturate the position and
orientation errors, $\|\tilde p\|\le\bar e_p$ and
$\angle(R_e^{\!\top}R_d)\le\bar e_R$, alongside the bound on the stiffness. Along a task direction $\hat t\in S^2$ and about a task axis $\hat a\in S^2$,
the nominal controller cannot exert more than
\begin{equation}
F_{\max} = \big(\hat t^{\!\top}\Lambda_{pp} K_{a,p}\,\hat t\big)\,\bar e_p ,
\quad
M_{\max} = \big(\hat a^{\!\top}\Lambda_{oo} K_{a,o}\,\hat a\big)\,\bar e_R ,
\label{eq:budget}
\end{equation}
where $\Lambda_{pp},\Lambda_{oo}$ and $K_{a,p},K_{a,o}$ are the translational
and rotational blocks of $\Lambda$ and $K_a$. The Cartesian inertia appears
because, without force feedback, the stiffness rendered towards the environment
is $K_c = \Lambda K_a$ rather than $K_a$: the gains in \eqref{eq:osc} are
accelerations, and the rendered stiffness is configuration dependent. Equation
\eqref{eq:budget} is therefore a local, quasi-static estimate, obtained by
neglecting the off-diagonal blocks of $\Lambda$. A resistance above it cannot
be overcome at the nominal stiffness, and raising the stiffness along the task
direction is the only remaining lever.

\subsection{Task direction from the policy intent}
\label{sec:direction}

Authority must be granted along the direction in which the task resists, and
the policy already states it: its action chunk is a short-horizon declaration
of intent. We extract two directions from it, a translational one $\hat t\in S^2$
and a rotational one $\hat a\in S^2$, and treat them as independent channels,
since a twist mixes linear and angular units and the two resistances are
physically distinct. In what follows, we describe the translational channel; the
rotational one is obtained by replacing $\hat t,\tilde p,\dot p_e,v_d$ with
$\hat a,e_O,\omega_e,\omega_d$.

We define the \emph{intent} as the chunk mean
$\bar v_d = \frac1H\sum_i v_d^{(i)}$, a forward-looking filter obtained for
free and far less noisy than a single streamed twist sampled from a multimodal
distribution. The task direction
is $\hat t^\star=\bar v_d/\|\bar v_d\|$, defined only when
$\|\bar v_d\|>v_{\min}$, since the direction of a vanishing vector is noise.


\subsection{Authority recognition}
\label{sec:detection}

To decide when authority is needed, the layer has to recognize whether the
policy is making progress on the task. The natural signals for this are the
commanded and the realized velocity along the task direction: the former states
what the policy is asking for, the latter what the robot is actually doing, and
their mismatch is a genuine error, which no force measurement can provide since
no desired force is available to compare against.
We therefore project the commanded and the realized motion on the task
direction, obtaining the commanded and realized rates of progress, 
\begin{equation}
\dot s_d = \hat t^{\!\top} v_d, \qquad
\dot s = \hat t^{\!\top} v ,
\label{eq:rates}
\end{equation}
and use them to
characterize the behaviour of the policy during a force-imposition task. Three
situations have to be distinguished: the robot is stuck against a resistance it
has to overcome ($\mathcal B$), the resistance is being overcome and the robot
advances ($\mathcal M$), and the policy no longer intends to move along that
direction ($\mathcal E$). Formally,
\begin{align}
\mathcal B :&\ \ \dot s_d>v_{\min}\ \wedge\ \dot s<\rho_{\mathrm{on}}\dot s_d
  \ \text{ for } T_{\mathrm{on}},
\label{eq:blocked}\\
\mathcal M :&\ \ \dot s_d>v_{\min}\ \wedge\ \dot s>\rho_{\mathrm{off}}\dot s_d,
\label{eq:moving}\\
\mathcal E :&\ \ \dot s_d<v_{\min} \ \text{ for } T_{\mathrm{off}},
\label{eq:ended}
\end{align}
with $0<\rho_{\mathrm{on}}<\rho_{\mathrm{off}}<1$ and $v_{\min}$ a threshold.
\begin{figure}
    \centering
    \vspace{12pt}
    \includegraphics[width=0.8\linewidth]{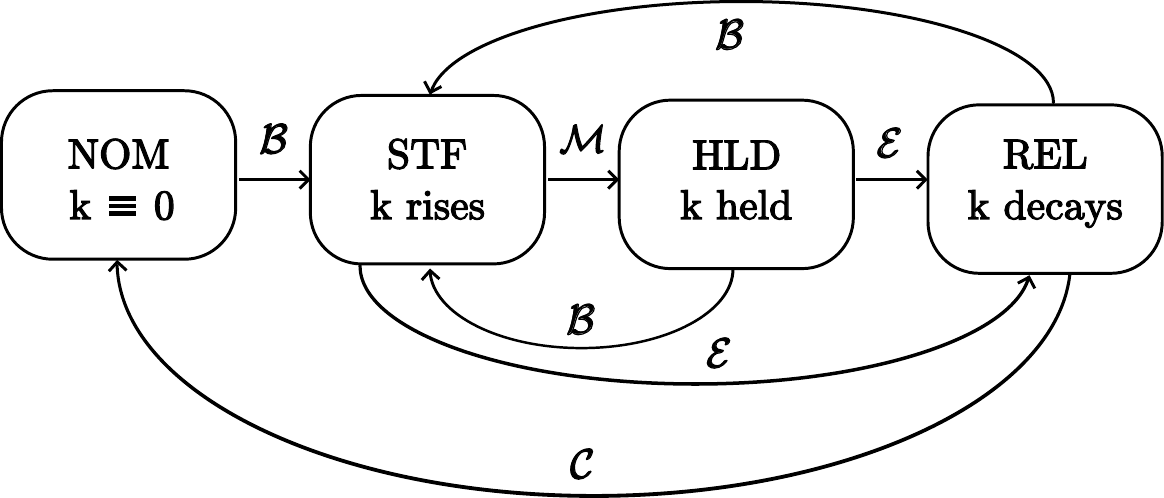}
    \caption{Supervisor state machine. Predicates $\mathcal B,\mathcal M,\mathcal E$
are defined in Sec.~\ref{sec:detection}, $\mathcal C$ in \eqref{eq:closing}.
}
    \label{fig:fsm}
    \vspace{-14pt}
\end{figure}

Predicate $\mathcal B$ thus holds when the policy is commanding a velocity but the robot
realizes less than a fraction $\rho_{\mathrm{on}}$ of what is commanded for a certain time $T_{\mathrm{on}}$.
Predicate $\mathcal M$ is its hysteretic counterpart, holding when the realized
speed exceeds a fraction $\rho_{\mathrm{off}}$ of the commanded one; the
gap between the two thresholds is what prevents chattering under stick--slip,
since a stall and its resolution are then separated by a band in which neither
predicate fires. Predicate $\mathcal E$ holds when the policy stops asking for
motion along the task direction, either because the pushing phase is over or
because it has turned elsewhere.


\subsection{Adaptation law}
\label{sec:supervisor}

Once a resistance is recognized, the layer has to decide how the added
stiffness $k$ evolves. A single continuous law is not sufficient. Granting
authority and withdrawing it are governed by different conditions, and the two
must be separated by hysteresis: if the stiffness were released as soon as the
robot starts moving, a residual resistance still above the ceiling
\eqref{eq:budget} would stall the robot again, and the adaptation itself would
produce a stick--slip limit cycle. Hysteresis requires memory of what the layer
is currently doing, so we formulate the adaptation as a hybrid automaton with
state $\sigma\in\{\textsc{Nom},\textsc{Stf},\textsc{Hld},\textsc{Rel}\}$, whose four modes have a direct physical reading:

\begin{itemize}
  \item \textsc{Nom}: the task progresses as demonstrated. The layer is
  inactive, $k\equiv0$, and the robot behaves exactly as during data
  collection. This is the only mode in which the task direction is free to
  follow the policy intent.
  \item \textsc{Stf}: a resistance has been recognized and the robot is not
  advancing. Authority is built up along the task direction.
  \item \textsc{Hld}: the robot is advancing again. The stiffness reached so
  far is retained, since the resistance is being overcome precisely because of
  it.
  \item \textsc{Rel}: the policy no longer asks for motion along that
  direction, so the authority is no longer needed and is returned to zero.
\end{itemize}
Transitions are driven by the predicates of Sec.~\ref{sec:detection}, plus the internal condition
\begin{equation}
\label{eq:closing}
    \mathcal C: k<\varepsilon\,k_{\max},
\end{equation}
 which closes the cycle once the added stiffness has decayed to a negligible fraction of its bound. The whole machine including states and transitions is shown in Fig.~\ref{fig:fsm}. Two of the transitions carry the adaptive behaviour: re-entering \textsc{Stf} from \textsc{Hld} handles a
resistance that reappears later along the task, with no friction model of any
kind, and re-entering it from \textsc{Rel} resumes from the current one rather
than from zero, so repeated stalls drive the stiffness towards the least
sufficient value.
 Note that \textsc{Stf} is left only on $\mathcal M$: if the policy stops
commanding motion while the robot is still blocked, the deficit $\delta^+$
vanishes and the stiffness stops growing, but it is retained until progress
resumes and the cycle continues through \textsc{Hld}. This is deliberate, since
the resistance is still present.
%
Within each mode the stiffness evolves as
\begin{equation}
\dot k =
\begin{cases}
\alpha\,\min\!\big(\lambda\delta^+,\,\dot k_{\max}\big)
  & \sigma=\textsc{Stf},\\[4pt]
-\bar\lambda\,k & \sigma=\textsc{Rel},\\[4pt]
0 & \text{otherwise,}
\end{cases}
\label{eq:kdot}
\end{equation}
with $\delta^+=\max(0,\dot s_d-\dot s)$ the
positive part of the velocity error, $\dot k_{\max}$ upper bound on stiffness time derivative, and $\lambda,\,\bar\lambda \in \mathbb{R}$ parameters which regulate the stiffness dynamics according to the criteria described in Sec.~\ref{sec:params}, and $\alpha\in\{0,1\}$ a gate discussed
in Sec.~\ref{sec:passivity}. The stiffness dynamics is then integrated and the result is mapped in the interval $[0,\,k_{\max}]$, with $k_{\max}$ upper bound on stiffness.
In \textsc{Stf} the rate
follows the error~\cite{yang2011humanlike}: the harder the policy pushes against a robot that does not
move, the faster authority is granted, and the growth stops of its own accord
as soon as the robot advances. 
Finally, the added stiffness is injected as a wrench acting along the task
direction alone,
\begin{equation}
\tau = \tau_0 + J^{\!\top}\mathcal W_{\!a},
\quad
\mathcal W_{\!a} =
\begin{pmatrix}
\big(k_t\,\hat t^{\!\top}\tilde p + d_t\,\hat t^{\!\top}\dot{\tilde p}\big)\hat t\\[2pt]
\big(k_a\,\hat a^{\!\top}e_O + d_a\,\hat a^{\!\top}\dot e_O\big)\hat a
\end{pmatrix},
\label{eq:wadd}
\end{equation}
with the added damping following the added stiffness,
$d_t=2\zeta\sqrt{m_t k_t}$ and $m_t=\hat t^{\!\top}\Lambda_{pp}\hat t$. Acting
through $J^{\!\top}$, rather than through an additional gain in
\eqref{eq:osc}, is what keeps the added force exactly along $\hat t$. Equation \eqref{eq:wadd} is therefore a rank-one modification of the rendered
Cartesian stiffness,
\begin{equation}
K_c \;\longrightarrow\; K_c + \Delta K_c,
\qquad
\Delta K_c = k_t\,\hat t\hat t^{\!\top} ,
\label{eq:rank1}
\end{equation}
with $k_t$ in N/m for the translational channel and $k_a$ in Nm/rad for the
rotational one,
that requires the task direction alone and no estimate of a full task frame,
and that leaves every direction orthogonal to $\hat t$ at its nominal
compliance: authority is raised without the off-axis forces of a uniformly
stiff robot.

\subsection{Energetic properties of the adaptation}
\label{sec:passivity}

A time-varying stiffness is not passive in general: the standard stability
analysis of impedance control does not hold once the impedance parameters vary,
and unstable behaviour can arise depending on how they are
varied~\cite{kronander2016stability}. Our layer varies the rendered stiffness in
two ways, in magnitude through $\dot k$ and in orientation through the task
direction, and both have to be accounted for.

Consider the storage function associated with the residual impedance,
\begin{equation}
V = \tfrac12\dot{\tilde x}^{\!\top}\Lambda\,\dot{\tilde x}
  + \tfrac12\tilde x^{\!\top}\big(K_c + \Delta K_c\big)\,\tilde x ,
\label{eq:storage}
\end{equation}
with $K_c = \Lambda K_a$ the stiffness rendered by the nominal controller and
$\Delta K_c$ the rank-one residual of \eqref{eq:rank1}. The analysis is carried
out per channel, so that $\Delta K_c = k\,\hat t\hat t^{\!\top}$ stands for
either of the two blocks; the two are coupled only through the shared tank
introduced below. Differentiating along the trajectories, and beyond the usual
dissipation and interaction terms, two contributions appear:
\begin{equation}
P_k = \tfrac12\,\dot k\,\big(\hat t^{\!\top}\tilde x\big)^2,
\qquad
P_{\mathrm{rot}} = k\,\big(\hat t^{\!\top}\tilde x\big)
                    \big(\dot{\hat t}^{\!\top}\tilde x\big).
\label{eq:pterms}
\end{equation}
The second is the power injected by \emph{reorienting} an anisotropic
stiffness, and it is sign indefinite: rotating $\hat t$ while $k>0$ changes the
stored energy with no motion of the robot at all. We remove it by construction,
freezing the task direction on leaving \textsc{Nom} and releasing it only on
return; since $k\equiv0$ in \textsc{Nom}, the direction may move only while the
residual is isotropic, in which case rotating it changes nothing, so
$P_{\mathrm{rot}}\equiv0$ throughout.

The first term is nonnegative while stiffening and nonpositive while releasing,
and we budget it with an energy tank~\cite{ferraguti2013tank}. Let $z$ be the
tank state, with energy $\psi = \tfrac12 z^2$ and $z(0)>0$, the latter avoiding
the singularity of
\begin{equation}
\dot z = \frac{\gamma}{z}\,\dot{\tilde x}^{\!\top}\hat D\,\dot{\tilde x}
       - \frac{\alpha}{z}\,\max\big(0,P_k\big)
\label{eq:tank}
\end{equation}
at the origin, so that the tank is filled by the power dissipated by the
damping and depleted by the power the stiffening injects. The two binary gates
\begin{equation}
\gamma = \mathbb 1\{\psi \le \psi_{\text{ub}}\},
\qquad
\alpha = \mathbb 1\{\psi \ge \psi_{\text{lb}}\},
\end{equation}
with $\psi_{\text{ub}}$ and $\psi_{\text{lb}}$ the upper and lower bounds on the stored
energy, prevent respectively an unbounded accumulation and a complete
depletion; $\alpha$ is the gate already appearing in \eqref{eq:kdot}, so that
stiffening is admitted only against energy previously dissipated, whereas
releasing, being passive, is always allowed. A single tank serves both
channels, so that the energy injected by the two is budgeted jointly.

The result we state concerns the residual layer rather than the closed loop as
a whole. Under exact model compensation, a nominal controller passive with
respect to the interaction port, positive definite nominal damping, a task
direction held constant while $k>0$, and a virtual equilibrium stationary
during the contact phase, the residual impedance gated by \eqref{eq:tank}
injects no net energy: substituting \eqref{eq:tank} into the derivative of
$\mathcal H = V + \psi$ and using $\gamma,\alpha\in\{0,1\}$ yields
$\dot{\mathcal H} \le \dot{\tilde x}^{\!\top}\mathcal F_{\mathrm{ext}}$.

The virtual equilibrium, being integrated from the policy twists, is treated
here as an exogenous input rather than as part of the analysis; its
contribution is bounded by the error saturation of Sec.~\ref{sec:impedance}.

\subsection{Parameters}
\label{sec:params}

Of the parameters appearing in Eqs.~\eqref{eq:blocked}--\eqref{eq:closing},
only $k_{\max}$ and $T_{\mathrm{ramp}}$ were tuned, once per platform and
channel; the remaining thresholds are either computed from the demonstrations
or fixed across both tasks. The first, $k_{\max}$, is set to the largest value the platform renders stably.
The second, $T_{\mathrm{ramp}}$, fixes how fast the stiffness may change, or
equivalently an
adaptation bandwidth $\bar\lambda=1/T_{\mathrm{ramp}}$, from which the
remaining gains follow as
\begin{equation}
\lambda = \frac{k_{\max}\bar\lambda}{v_{\mathrm{ref}}},
\qquad
\dot k_{\max} = k_{\max}\bar\lambda .
\label{eq:gains}
\end{equation}
The same bandwidth governs the release, which decays with time constant
$1/\bar\lambda$.
The stiffness then grows by at most a fraction $\bar\lambda$ of its bound per
second, so $T_{\mathrm{ramp}}$ is the minimum time to reach $k_{\max}$,
attained when the robot is fully blocked and the policy pushes at the reference
speed $v_{\mathrm{ref}}$; slower commands give a slower, safer ramp. We choose it well above the time scale of the stiffened mechanics,
$T_{\mathrm{ramp}}\ge10\sqrt{m_t/k_{\max}}$, which separates the adaptation
from the dynamics it adapts.

Two quantities come from the demonstrations instead of a design choice:
$v_{\mathrm{ref}}$, the $95$th percentile of the commanded speed projected on
the task direction over the frames in which the policy is pushing, and
$v_{\min}$, $20\%$ of the median intent magnitude $\|\bar v_d\|$ over the same
demonstrations. The rest are
dimensionless constants, identical across tasks and channels: a stall is
declared below a quarter of the commanded speed and resolved above one half
($\rho_{\mathrm{on}}=0.25$, $\rho_{\mathrm{off}}=0.5$), the dwell times are one
and three action chunks ($T_{\mathrm{on}}=H/f_s$, $T_{\mathrm{off}}=3H/f_s$),
and the added damper and the closing threshold are $\zeta=0.7$,
$\varepsilon=0.01$.

\section{Experimental evaluation}
\label{sec:experiments}

The evaluation is designed to answer two questions. First, whether the proposed layer recovers the failures of the policy without retraining it, and whether it does so by discovering the required authority at run time, having
seen neither the resistance nor any force measurement. Second, whether the
anisotropy of the layer matters: a uniformly stiff robot would overcome the
same resistance, so we ask what that costs in terms of the forces produced
along the directions the task does not require.

\subsection{Experimental setup}

Experiments are performed on a Franka Research 3 (FR3) torque-controlled at
1~kHz, with a single wrist-mounted Intel RealSense D455 providing
$224\times224$ RGB images. Demonstrations are collected by teleoperation
through a Haption Virtuose 6D Desktop force-feedback device at $20$~Hz. Each
demonstration records the end-effector pose and the image as
the observation, and the commanded twist as the action.
Two tasks instantiate two classes of resistance. 
%

\emph{Unscrewing} requires torque about the axis of a bottle: the robot, already grasping
the hexagonal cap, unscrews it against a breakaway torque set by how tightly
the cap was closed. The task succeeds when the cap has rotated by $180^\circ$
relative to its initial pose. Only the rotational channel of the layer is enabled. To hold the bottle, we use another robot controlled independently.

\emph{Pushing} requires force along a rail: the robot pushes a ball-spring
T-nut along the T-slot groove of an aluminium extrusion and, at a point unknown
to the policy, meets a number of further T-nuts placed in series whose friction
it must overcome and keep overcoming for the rest of the stroke. The task
succeeds when the T-nut has advanced $18.5$~cm past its initial position. Only
the translational channel is enabled.

Which channel is enabled, whether linear or rotational, is the only task-specific statement made to the layer;
the direction within the channel is estimated online.

For each task we prepare six discrete levels of resistance (L1 to L6), obtained by
varying the thread tightness of the cap and the number of T-nuts in series respectively. The \emph{first four} levels are used for data
collection, with $15$ demonstrations per level for unscrewing ($60$ in total)
and $12$ per level for pushing ($48$ in total). The last two levels are
withheld entirely from the demonstrations and reserved for evaluation. Last two levels require higher force or torque, respectively, to be accomplished, with respect to the other ones.
The six levels are characterized \emph{a posteriori} by the median peak force
the proposed layer exerts along the task direction over its successful trials,
reported in Table~\ref{tab:levels}: the two withheld levels demand roughly
twice the torque of the demonstrated ones in unscrewing, and in pushing the
demanded force grows monotonically from $17$ to $63$~N.

\begin{table}[b]
\centering
\caption{Authority demanded by each level.}
\label{tab:levels}
\footnotesize
\begin{tabular}{lcccccc}
\toprule
 & L1 & L2 & L3 & L4 & L5 & L6 \\
\midrule
Unscrewing [Nm] & $1.42$ & $1.58$ & $1.36$ & $1.54$ & $2.19$ & $2.76$ \\
Pushing [N] & $17.2$ & $26.5$ & $31.3$ & $41.2$ & $51.1$ & $63.2$ \\
\bottomrule
\end{tabular}
\end{table}

On this dataset we train a flow-matching visuomotor policy: a conditional
U-Net, conditioned on the encoded visual and proprioceptive observation,
predicts the velocity field of a rectified flow between Gaussian noise and the
demonstrated action chunk, integrated at inference time with an Euler solver.
The same checkpoint is used in every condition below, with no retraining and
no change to its inputs or outputs.

\subsection{Evaluation protocol}

Evaluation covers all \emph{six} levels: the four seen during data collection
and the two withheld ones. Each level is repeated five times under each
condition. In the pushing task the point at which the resistance
appears is randomized at every trial, as it was during demonstration. A trial
is declared failed if the success criterion is not met before a fixed timeout,
identical across conditions.

Three conditions are compared, all executing the same policy checkpoint:
\begin{itemize}
  \item Fixed $K_0$: the baseline impedance controller, with the gains
  used to collect the demonstrations and the proposed layer disabled. 
  \item Proposed: the same controller with the stiffening layer
  active underneath the same, unmodified policy.
  \item Fixed $K_{\mathrm{high}}$: the baseline with uniformly raised stiffness, run at the highest resistance level as a reference for what
  isotropic stiffening costs.
\end{itemize}
Table~\ref{tab:params} lists the parameters of the policy and of the layer for
both tasks. In every trial we also measure the interaction forces, used for
analysis only: the layer consumes none of them at run time.

\begin{table}[t]
\vspace{5pt}
\centering
\caption{Parameters of the policy and of the stiffening layer. 
}
\label{tab:params}
\footnotesize
\begin{tabular}{lcc}
\toprule
 & Unscrewing & Pushing \\
\midrule
\multicolumn{3}{l}{\emph{Policy}}\\
chunk length $H$ / rate $f_s$ & \multicolumn{2}{c}{8 / 20~Hz}\\
demonstrations per level & 15 & 12 \\
\midrule
\multicolumn{3}{l}{\emph{Baseline controller}}\\
$K_{a,p}$ [s$^{-2}$] & 500 & 300 \\
$K_{a,o}$ [s$^{-2}$] & 250 & 600 \\
$\bar e_p$ [m] / $\bar e_R$ [rad] & \multicolumn{2}{c}{0.05 / 1.0}\\
\midrule
\multicolumn{3}{l}{\emph{Stiffening layer}}\\
active channel & rotational & translational \\
$k_{\max}$ & 500~Nm/rad & 2000~N/m \\
$v_{\min}$ & 0.0051~rad/s & 0.0019~m/s \\
$v_{\mathrm{ref}}$ & 0.093~rad/s & 0.026~m/s \\
$T_{\mathrm{ramp}}$ [s] & \multicolumn{2}{c}{7.5}\\
$T_{\mathrm{on}}$ / $T_{\mathrm{off}}$ [s] & \multicolumn{2}{c}{0.4 / 1.2}\\
$\rho_{\mathrm{on}}$, $\rho_{\mathrm{off}}$, $\zeta$, $\varepsilon$
  & \multicolumn{2}{c}{0.25, 0.5, 0.7, 0.01}\\
\bottomrule
\end{tabular}
\vspace{-10pt}
\end{table}

\subsection{Results}

The layer introduced above improves the success rate on both tasks (Table~\ref{tab:success}). In
unscrewing the baseline solves every demonstrated level and none of the two
withheld ones, while the layer solves the first of them in every trial and the
second in three out of five. In pushing the improvement is broader, because
there the baseline already breaks down at two of the demonstrated levels: the layer recovers those as well, and solves eight of the ten trials at the withheld levels. This is worth stating separately, since the layer is not only a
mechanism for extrapolation: where the resistance is unpredictable, as when the T-nut meets a variable number of others at a random point, the nominal controller is already insufficient \emph{within} the demonstrated range. With five trials per cell the individual figures carry substantial uncertainty:
a success of $3/5$ has a $95\%$ exact binomial interval of $[15\%, 95\%]$. The
comparison therefore rests on the pattern across levels rather than on single
cells.
The detection and the actuation can be separated, because the layer runs in
both conditions and only the added wrench is suppressed under the baseline. In $28$ of the $29$ failed baseline trials the supervisor did recognize the stall, at a median of $16.8$~s into the trial: what
the baseline lacks is therefore not the ability to detect that it is being
resisted, but the authority to act on it.


\begin{table}[t]
\vspace{5pt}
\centering
\caption{Success rate [\%] over five trials per level. L1--L4 were demonstrated,
L5--L6 withheld.}
\label{tab:success}
\footnotesize
\begin{tabular}{llcccccc}
\toprule
Task & Condition & L1 & L2 & L3 & L4 & L5 & L6 \\
\midrule
\multirow{2}{*}{Unscrewing}
 & $K_0$ & $100$ & $100$ & $100$ & $100$ & $0$ & $0$ \\
 & Prop. & $100$ & $100$ & $100$ & $100$ & $100$ & $60$ \\
\midrule
\multirow{2}{*}{Pushing}
 & $K_0$ & $100$ & $100$ & $20$ & $0$ & $0$ & $0$ \\
 & Prop. & $100$ & $100$ & $80$ & $80$ & $100$ & $60$ \\
\bottomrule
\end{tabular}
\vspace{-10pt}
\end{table}




\begin{figure*}[t]
\vspace{5pt}
\centering
  \includegraphics[width=0.35\linewidth]{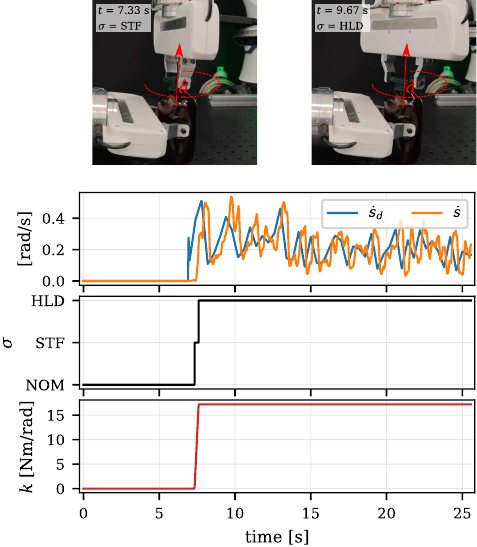}
  \hspace{3cm}
  \includegraphics[width=0.35\linewidth]{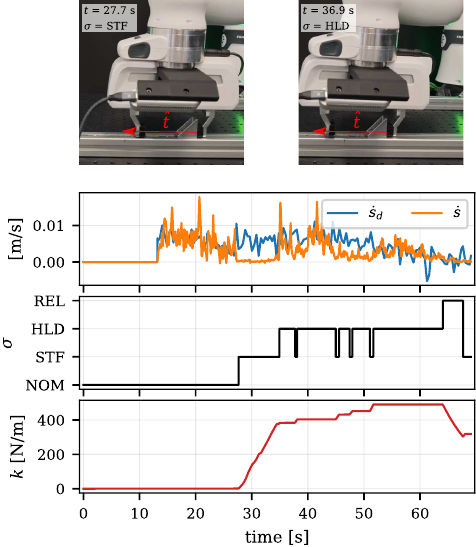}
\caption{Representative trials at the highest resistance level, unscrewing
(left) and pushing (right). The snapshots are taken as the supervisor enters
\textsc{Stf}, and two seconds after it enters \textsc{Hld}, having overcome the resistance, and report the
recovered task direction, the mode and the elapsed time. Below: commanded and realized rates of
progress $\dot s_d,\dot s$; supervisor mode $\sigma$; added stiffness $k$.}
\label{fig:trials}
\vspace{-14pt}
\end{figure*}

Fig.~\ref{fig:trials} shows how this happens in a single trial of each task. The stall
appears as a persistent gap between the commanded and the realized rate of
progress, the policy continuing to command motion that the robot does not
execute; this is the only signal the layer uses. The supervisor confirms it after one action chunk, freezes the
task direction and begins to build authority, until motion resumes and the mode
moves to \textsc{Hld}. In pushing the mode alternates between \textsc{Stf} and \textsc{Hld}
several times as the T-nut advances, each re-entry resuming from the current
stiffness, so that authority accumulates only as far as the task demands.

The stiffness the layer settles on reflects that demand: in pushing the added
stiffness grows with the level, averaging about $100$~N/m at the lowest and
$600$~N/m at the highest, while remaining well below the bound of $2000$~N/m.

The layer has a cost in execution time: in pushing the median completion time
of a successful trial grows from $45.4$ to $64.8$~s, the difference being the
dwell before a stall is declared and the deliberately slow ramp, whereas in
unscrewing it is slightly lower, $26.6$ against $29.6$~s, because the authority
supplied shortens the stall. 

\subsection{Comparison with a uniformly stiff baseline}

An arbitrarily isotropic stiff controller would solve the success-rate problem,
at the price of interaction forces larger than the task requires. This is not a
matter of tuning: neither the magnitude nor the direction of the stiffness can
be chosen in advance. The magnitude depends on friction conditions that are
unknown before the contact and vary from one instance to the next, and the
direction is whichever way the policy happens to be pushing at that moment. Any
fixed, isotropic choice is therefore either insufficient for the resistance encountered, or needlessly stiff everywhere else.

Table~\ref{tab:offaxis} quantifies the difference on the pushing task at level
6, over 5 trials, where the forces involved make it most apparent: the uniformly stiff controller produces a median of $47.5$~N orthogonal to the task direction against $18.3$~N for the proposed layer, and the off-axis to task force ratio drops from $0.63$ to $0.30$. The spread is also much wider: across the five trials the uniformly stiff controller reaches up to $107$~N off-axis, against $21$~N for the layer. Granting authority only along the direction the policy is pushing, and only once the robot is seen not to advance, delivers the force the task requires without exceeding it elsewhere: this is the primary advantage offered by the rank-one construction in~\eqref{eq:rank1}.

\begin{table}[t]
\vspace{5pt}
\centering
\caption{Peak force along and orthogonal to the task direction, pushing at L6.
Median over five successful trials, range in brackets.}
\label{tab:offaxis}
\footnotesize
\begin{tabular}{lccc}
\toprule
Condition & Along task [N] & Off-axis [N] & Ratio \\
\midrule
Proposed & $62.0$ \footnotesize{[58.0, 64.5]} & $18.3$ \footnotesize{[13.9, 21.0]} & $0.30$ \\
Fixed $K_{\mathrm{high}}$ & $75.2$ \footnotesize{[64.2, 112.7]} & $47.5$ \footnotesize{[30.8, 106.6]} & $0.63$ \\
\bottomrule
\end{tabular}
\vspace{-10pt}
\end{table}

\subsection{Energy budget}

The tank never constrained the adaptation: starting from $3$~J with a lower
bound of $0.1$~J, the lowest level reached over all trials was $2.86$~J in
unscrewing and $1.88$~J in pushing, so the gate never suspended the stiffening.
The margin follows from the slow ramp, which injects little power at any
instant. The guarantee of Sec.~\ref{sec:passivity} is thus a property of the
design rather than an active constraint here.

\subsection{Limitations}

Our framework is specific for tasks in which a resistance has to be overcome. For this reason, any other task in which it is necessary to increase compliance when contact forces rise, or tasks which can include interaction with unmodeled
obstacles, is not supported in its current form. A natural future development would be to integrate a higher level reasoning that determines whether the task requires imposing force or requires compliance for safety.

\section{Conclusion}
\label{sec:conclusion}

In this work, we addressed a critical failure mode of generative visuomotor policies in force-imposition tasks: when unobserved physical resistance exceeds the nominal capabilities of the underlying impedance controller, tasks fail despite accurate policy execution. 
To resolve this, we introduced a residual stiffening layer positioned beneath the policy. 
Our framework detects resistance from the tracking error between commanded and actual robot states. 
A four-mode supervisor dynamically scales stiffness exclusively along the policy’s inferred directional intent, while a directional constraint and an energy-tank gate prevent unstable energy injection into the system interaction.

We showed that on two real tasks, unscrewing a bottle cap and pushing a T-nut against non-homogeneous friction, the layer recovers the failures of the fixed-compliance
baseline at resistance levels never demonstrated, and also at demonstrated
levels where the resistance is unpredictable and the nominal controller is  insufficient. 
A uniformly stiff controller, by contrast, cannot know in advance either the magnitude or the direction of the required stiffness, and therefore produces large undesired forces orthogonal to the task direction.
\bibliographystyle{ieeetr}
\bibliography{bibliography}

\end{document}